%% file: acl_latex.tex
\documentclass[11pt]{article}

\usepackage[final]{acl}

\usepackage{times}
\usepackage{latexsym}

\usepackage[T1]{fontenc}

\usepackage[utf8]{inputenc}

\usepackage{microtype}

\usepackage{inconsolata}

\usepackage{graphicx}
\usepackage{amsmath}
\usepackage{xcolor}
\usepackage{booktabs}
\usepackage{multirow}
\usepackage{placeins}
\usepackage{float}
\usepackage{enumitem}
\usepackage[most]{tcolorbox}
\tcbuselibrary{breakable,listings,skins}
\usepackage{listings}
\lstdefinelanguage{prompt}{
  basicstyle=\scriptsize\ttfamily,
  moredelim=[s][\color{gray}]{\{}{\}},
}

\newtcblisting[use counter=promptcounter]{prompt}[1][]{
  enhanced,
  arc=0em,
  boxrule=0.5pt,
  listing only,
  listing options={
    numbers=none,
    frame=none,
    language=prompt,
    showstringspaces=false,
    showspaces=false,
    showtabs=false,
    basicstyle=\ttfamily\scriptsize,
    upquote=true,
    breaklines=true,
    breakindent=0pt,
    columns=fullflexible,
  },
  top=2pt,
  bottom=2pt,
  left=2pt,
  right=2pt,
  colback=white,
  colframe=gray,
}

\definecolor{revcolor}{RGB}{255,0,0}

\newcommand{\reffig}[1]{Figure~\ref{#1}}
\title{Inducing Task Models from Computer-Use Traces}

\author{Yucheng Jiang$^{1}$ \quad Zora Zhiruo Wang$^{2}$ \quad Ruishi Chen$^{1}$ \quad Diyi Yang$^{1}$ \\
  $^{1}$Stanford University \quad $^{2}$Carnegie Mellon University \\
  \texttt{\{yuchengj,diyiy\}@cs.stanford.edu}}

\begin{document}
\maketitle
\begin{abstract}
\input{sections/0_abstract}
\end{abstract}

\section{Introduction}
\label{sec:introduction}
\input{sections/1_introduction}

\section{Problem Formulation}
\label{sec:formulation}
\input{sections/2_task_model_induction}

\section{Method}
\label{sec:method}
\input{sections/3_method}

\section{Intrinsic Evaluation}
\label{sec:experiment}
\input{sections/4_intrinsic_evaluation}

\section{Extrinsic Evaluation}
\input{sections/5_extrinsic_evaluation}

\section{Related Work}
\label{sec:related-works}
\input{sections/6_related_works}

\section{Conclusion}
\label{sec:conclusion}
\input{sections/7_conclusion}

\section*{Limitations}
\input{sections/9_limitation}

\section*{Ethics Statement}
\input{sections/8_ethics}

\section*{Acknowledgments}
We thank Harshit Joshi, Vishakh Padmakumar, Michael Ryan, Jiacheng Sang, Yijia Shao, Yilin Xu, John Yang, Ruozhen Yang, Dora Zhao, Cyrus Zhou, and Ziran Zhou for their thoughtful feedback, discussions, and support throughout the project. This work is supported in part by grant from Laude Moonshot Seed Grant, a Stanford HAI-Banco Itau collaboration, and ONR N000142412532. 

\bibliography{custom}

\clearpage
\appendix
\label{sec:appendix}
\input{appendix/structural_validity}
\input{appendix/latent_task_induction_robustness}
\input{appendix/skillsbench_trajectory_conversion}

\input{appendix/task_model_fidelity_rubric}
\input{appendix/objective_rubric_full_results}
\input{appendix/pipeline_prompts}

\input{appendix/worked_example}
\end{document}

%% file: sections/0_abstract.tex
Naturalistic computer-use traces, passively recorded screenshots and mouse or keyboard actions, are a valuable resource for deriving symbolic, auditable, and reusable models of how everyday work is done.
Such models matter as computer-use agents enter real work, where agents need to learn how tasks are actually performed, and organizations need to audit and reuse that knowledge. However, inducing such task models is challenging, as activity is observed only as low-level events and real-world work is multi-threaded with interleaved goals. Existing methods assume a given task or a single workflow, and produce step-level summaries rather than structured task models.
We introduce \textsc{Task Model Induction} (TMI), which (i) discovers the latent tasks in an unconstrained trace, disentangling concurrent activity, and (ii) for each latent task, induces a task model pairing a hierarchical objective model of recursive goal decomposition with a procedure model of the control flow that organized the execution. Intrinsically, on controlled human and agent trajectories, TMI recovers interleaved tasks with 0.974 agreement against ground-truth groupings and reconstructs 74.9\% of the observed execution steps, far more than the strongest workflow induction baseline. Extrinsically, skills derived from TMI's task models improve held-out task accuracy by 30.0\% over the strongest baseline. \footnote{Our codebase is available at \url{https://github.com/Yucheng-Jiang/task-model-induction}.}

%% file: sections/1_introduction.tex
Naturalistic computer-use activity traces, passively recorded sequences of screenshots and mouse or keyboard events, are a valuable resource for deriving symbolic, auditable, and reusable models of how everyday work is done~\cite{shaikh2025gum,wang2025ai}. Most of this work is never documented, and the expertise it encodes remains tacit. Recovering the objectives a user pursued and the procedures they followed turns a recording into an explicit account of what was accomplished and how, which people can audit for systematic patterns in human and AI work~\cite{wang2025ai}, reuse as documentation, and transfer to new practitioners. Beyond human use, such traces support agent learning from human activity without costly annotation~\cite{lu2025video,song2026watchlearnlearninguse}, and personalization, where activity-based user models let systems anticipate user goals~\cite{shaikh2026learning}.

\begin{figure}[t]
    \centering
    \includegraphics[width=\linewidth]{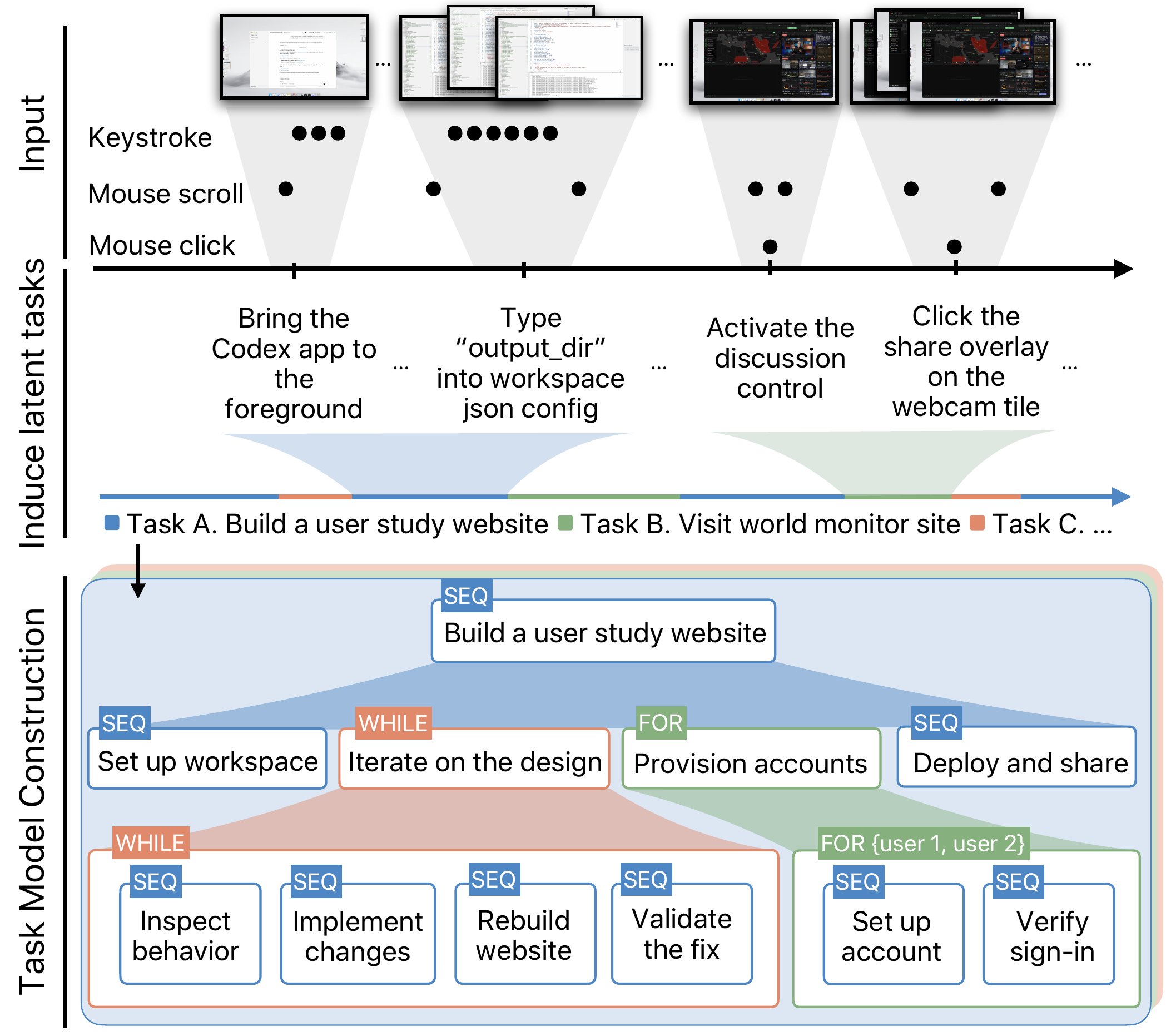}
    \caption{\textbf{Task model induction from computer-use traces.} Screenshots and input events are grounded into semantic actions, whose interleaved spans are assigned to latent tasks. Each task is represented as a hierarchy pairing objectives with control-flow operators.
    }
    \label{fig:teaser}
    \vspace{-0.75em}
\end{figure}

Modeling these traces, however, poses three challenges. At the \emph{signal} level, raw events such as cursor movements, key presses, and screen pixels carry little semantic meaning, and bridging perceptual observations to semantic intent requires substantial interpretation~\cite{yang2023setofmark,you2024ferret,zheng2024gpt,shaw2023pixels}. At the \emph{structural} level, naturalistic work is multi-threaded, with users switching among unrelated goals within a single session and pursuing interleaved sub-objectives within a single task~\cite{czerwinski2004diary,gonzalez2004constant,mark2008cost,adamczyk2004if,iqbal2007task,salvucci2009threaded}. At the \emph{representational} level, a trace captures the executed path but does not directly reveal either the goal hierarchy that motivated each step or the control flow that organized the execution~\cite{stanton2006hierarchical,card2018psychology,diaper2003handbook}.

Prior approaches address these challenges only partially. LLM summarization condenses the session into free-form prose, blending concurrent tasks and discarding goal hierarchy and control flow. Prompting an LLM directly for a task model recovers the schema but not the execution structure, as our direct generation baseline shows (\S\ref{sec:structural_validity}).
Workflow induction recovers a stepwise decomposition of the activity~\cite{chen2024sharinganextractuseraction,wang2025ai} but treats the recording as one continuous workflow, leaving interleaved tasks entangled, sub-objectives flattened, and iteration and branching unexpressed~\cite{wang2025ai,zang2025ms4ui}. Trace-analysis methods assume the root task is given in advance~\cite{wang2024agent,grohs2024beyond} and do not extend to sessions with latent and diverse tasks.

We address these gaps by defining a \textbf{task model}, an explicit representation of a single task that pairs its goal hierarchy with the control flow of its execution. The \emph{objective model} represents the goal hierarchy as a recursive decomposition of the task into the (sub-)objectives the user pursued~\cite{wing2006Computation,stanton2006hierarchical}. The \emph{procedure model} represents the control flow as a composition of sequencing and iteration operators, the structured programming constructs that remain observable in a trace~\cite{bohm1966flow}. Recovering task models from naturalistic computer-use traces (\S\ref{sec:formulation}) requires a system to jointly discover the latent tasks of an unconstrained session and induce a task model for each, with no tasks, boundaries, or descriptions given in advance (\reffig{fig:teaser}).

We introduce \textsc{Task Model Induction} (TMI) (\S\ref{sec:method}), a method that addresses the three challenges in turn. \emph{Event grounding and activity segmentation} recovers what each raw event did from its visual context and groups the results into semantic actions and activities, bridging low-level signal and local intent (\S\ref{sec:activity-abstraction}). \emph{Latent task induction} untangles the multi-threaded activity stream, discovering tasks and assigning possibly non-contiguous activities to each without a pre-specified task set (\S\ref{sec:latent-task-induction}). \emph{Task model construction} builds each task's objective and procedure models under formal validity constraints and reconciles them into a unified model in which objective scope and control flow are mutually consistent (\S\ref{sec:task-model-construction}).

We evaluate TMI both intrinsically via controlled reconstructions of human computer-use sessions (\S\ref{sec:experiment}), and extrinsically via downstream agent learning (\S\ref{sec:extrinsic}). On trajectories built from a dataset of recorded real human work sessions~\citep{wang2025ai}, our method recovers interleaved tasks with 0.974 agreement against ground-truth task groupings and matches 74.9\% of observed execution steps against 30.3\% for the strongest workflow induction baseline.
Using the induced task model to generate reusable agent skills, evaluated on held-out SkillLearnBench~\citep{zhong2026skilllearnbench} tasks, improves task accuracy by 30.0\% over baseline.

Our contributions are as follows.
\begin{itemize}
    \item We formalize the \textbf{task model} for induction from unconstrained traces, a representation that pairs hierarchical objective decomposition with structured control flow, grounded in computational thinking and the structured programming theorem.
    \item We propose \textbf{Task Model Induction (TMI)}, a method that grounds raw events into semantic activities, untangles interleaved sessions into latent tasks, and induces each task's objective and procedure models independently before reconciling them into a unified model under formal validity constraints.
    \item Intrinsic and extrinsic evaluation show that our method recovers interleaved tasks and their execution structure more faithfully than workflow induction baselines, and that the induced task models yield more effective skills for downstream agents.
\end{itemize}

%% file: sections/2_task_model_induction.tex
Describing the activity conducted in a computer-use session is necessary for analyzing, learning, and auditing it. We formalize this process as the recovery, from a naturalistic computer-use trace, of the latent tasks and their task models specifying the objectives and procedures.

Let $X = \langle x_1, x_2, \ldots, x_N \rangle$ be a user's computer-use trace in an interactive computer environment, where each event $x_i = (s_i, a_i, \tau_i)$ contains the state represented as a screenshot $s_i$, a low-level operation (e.g., \texttt{click}), and a timestamp $\tau_i$.

The trajectory reflects the user pursuing a set of latent tasks $\mathcal{T} = \{t_1, t_2, \ldots\}$, where a possibly non-contiguous subsequence of events collectively realizes a task $t_j$. However, inducing these latent tasks remains an open challenge, as none of the tasks, their boundaries, or the event-to-task assignments are predefined in realistic human traces.

We define the induction problem as jointly discovering the set of latent tasks $\mathcal{T}$ and, for each task $t \in \mathcal{T}$, inducing a task model $M_t$ that unifies two complementary axes.
\begin{itemize}
\item The \emph{objective model} $O_t$ is a hierarchical decomposition of the objective of $t$ into the sub-objectives the user pursues.
\item The \emph{procedure model} $P_t$ is a trace-grounded composition of control-flow operators specifying how the execution of $t$ is organized through sequencing and iteration.
\end{itemize}
The task model $M_t$ is structured as a tree, whose every node pairs an objective with a control-flow operator over its children.

%% file: sections/3_method.tex
\begin{figure*}[t]
    \centering
    \includegraphics[width=\textwidth]{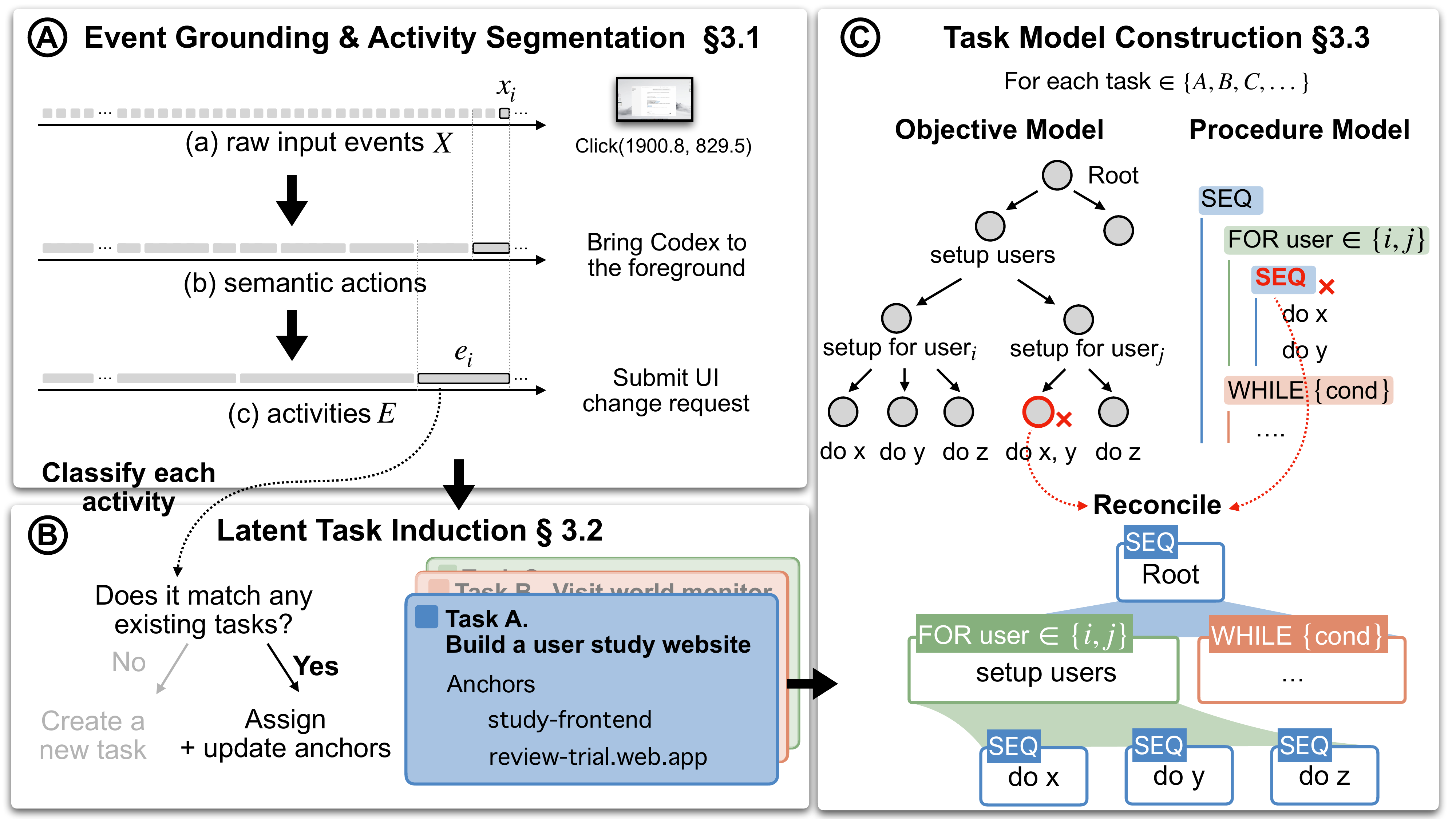}
    \caption{\textbf{Overview of \textsc{Task Model Induction} (TMI).} \emph{(1) Event grounding and activity segmentation} reads each mouse and keyboard event in $X$ against its visual context and groups the results into semantic actions and then activities $E$ (\S\ref{sec:activity-abstraction}). \emph{(2) Latent task induction} assigns each activity to the closest existing task or opens a new one, with identifiers keeping a task intact across application and naming shifts (\S\ref{sec:latent-task-induction}). \emph{(3) Task model construction} induces an objective model and a procedure model for each discovered task and reconciles them into one task model whose nodes carry both an objective and a control-flow operator (\S\ref{sec:task-model-construction}). Red marks in stage 3 show what each model misses on its own, \emph{do x} and \emph{do y} collapsed into one leaf by the objective model and \emph{do z} dropped from the loop body by the procedure model, both recovered in the reconciled \emph{for-each}. \reffig{fig:worked-example} shows the tasks and the task model induced from one recorded session.
    }
    \label{fig:method}
\end{figure*}

TMI recovers task models from raw computer-use traces in three stages (\reffig{fig:method}). \emph{Event grounding and activity segmentation} recovers what each event did from its visual context and groups the grounded events into semantic actions and activities (\S\ref{sec:activity-abstraction}). \emph{Latent task induction} discovers the tasks pursued in the session and assigns each activity to the task it realizes (\S\ref{sec:latent-task-induction}). \emph{Task model construction} builds complementary objective and procedure models for each task and reconciles them into a single task model, in which every node pairs an objective with the control-flow operator over its children (\S\ref{sec:task-model-construction}).

\subsection{Event Grounding and Activity Segmentation}\label{sec:activity-abstraction}
While a raw event (e.g., \texttt{click(1900.8, 29.5)}) carries no intent on its own, the trace can be abstracted into semantic units that support meaningful inference of user intent. Each event is grounded in the screenshots that bracket it, which supply the visual context the event alone lacks, and the grounded trace is segmented into semantic actions and then \emph{activities} (\reffig{fig:method}(a)), the atomic unit at which user intent can be inferred~\citep{leont2024problem}.

Grounding infers the meaning of an event from the visual change it produced. A vision-language model receives the screenshot pair $(s_i, s_{i+1})$ bracketing each event $x_i$ together with the recorded operation $a_i$, and reports what was done, to which artifact, in which application, along with an OCR transcript of the visible screen. The difference between the two screenshots indicates what the event altered (e.g., \texttt{click(1900.8, 29.5)} to \emph{bring the Codex app to the foreground}). Grounding stays constrained to evidence visible in the screenshots, consulting $s_{i+1}$ only to clarify what changed, not to impute retroactive intent.

Segmentation groups grounded events into units of work at two layers of abstractions. First, a \emph{semantic action} collects the consecutive events that together produce one meaningful state change in an artifact (e.g., \emph{edit \mbox{\texttt{app/page.tsx}} to revise the consent form copy}). Next, an \emph{activity} collects the semantic actions that one local objective explains (e.g., \emph{submit a UI change request}), beginning when that objective is adopted and ending when it is achieved, abandoned, or superseded. The result is an activity sequence $E = \langle e_1, e_2, \ldots, e_K \rangle$.

A language model (\S\ref{sec:implementation}) segments each of the two levels once, in the direction where the evidence for that level's boundaries lies~\citep{zacks2001event}. Semantic actions are segmented backward over the grounded events, since a semantic action ends where an artifact reaches its new state and only the events that follow confirm that the state was reached. Activities are segmented forward along the resulting semantic actions, since an activity begins when its objective is adopted, and only the preceding context signals that adoption.

\subsection{Latent Task Induction}\label{sec:latent-task-induction}
Naturalistic activity is multi-threaded (\reffig{fig:method}(b)), so the activities of one task are interleaved with those of others and spread across applications and artifacts~\citep{czerwinski2004diary,gonzalez2004constant}. Latent task induction recovers the task set $\mathcal{T}$, in which every activity belongs to exactly one task and each task states the objective its activities jointly support. Activities are processed in the order they appear in the trace, and each is assigned to a task. The tasks discovered this way are consolidated once the trace ends.

Assigning an activity requires a compact representation of each task that the activity can be compared against. Each task maintains a profile with a summary of what the task achieves and a small set of referential identifiers, such as artifacts and named entities that recur across its activities (\reffig{fig:method}(b)). $\mathcal{T}$ starts empty and grows as each activity $e_k$ is compared against these profiles, joining the semantically closest task $t \in \mathcal{T}$ or opening a new task when none subsumes it. The profile is updated as activities are assigned, so its summary tracks the task's evolving scope and its identifiers accumulate the aliases under which the task appears.

Identifiers hold a task together under surface variation. A single task commonly spans multiple applications, artifacts, and referential aliases (e.g., a user-study website appears as the repository \texttt{study-frontend} and the deployment URL \texttt{review-trial.web.app}).
The summary drifts under this variation, whereas the identifiers cross-reference a task across changes in application, artifact, and naming, so
a coherent task is not fragmented into spurious subtasks.

Incremental assignment alone splits a task when its objective drifts or when related activities are separated by long interruptions. A global consolidation pass therefore examines all discovered tasks and merges those that pursue the same objective.

\subsection{Task Model Construction}\label{sec:task-model-construction}
Modeling a task $t$ requires understanding the two types of evidence it carries, procedure and objective.
Constructing a task model thus means inducing an objective model $O_t$ and a procedure model $P_t$ from the activities $E_t$ of each discovered task $t$, then reconciling them into the task model $M_t$ (\reffig{fig:method}(c)). Each is induced independently, so that its structure is resolved under its own evidence before being constrained by the other.

\paragraph{Objective model.}
The objective model is the hierarchy that explains why the observed activities were performed. Its root is the task objective, its leaves are the activities $E_t$, and its internal nodes are latent sub-objectives that jointly explain their descendants. A language model induces $O_t$ from the task objective and $E_t$ by recursive decomposition~\citep{wing2006Computation}, breaking a goal that cannot be pursued directly into sub-goals whose union covers the relevant evidence.

Each node denotes the outcome to be achieved, abstracting over the particular strategy used to reach it (e.g., \emph{provision reviewer accounts} instead of \emph{set up each account and verify its sign-in}). The children of a node are necessary components or preconditions of their parent and jointly account for its observed activities. A node spanning many activities is refined further (e.g., \emph{iterate on the design} decomposes into \emph{implement changes} and \emph{validate the fix}), whereas a node grounded in a single activity has reached the level of a local objective and remains a leaf. The resulting hierarchy is abstract enough to transfer across executions yet grounded enough that every leaf stays tied to an observed activity. Objective-side validity constraints are detailed in Appendix~\ref{app:structural-validity}.

\paragraph{Procedure model.}
The procedure model captures how the execution of the task was organized in time. Following the structured programming theorem~\citep{bohm1966flow}, any procedure can be expressed with sequencing, selection, and iteration. In our setting, however, selection is usually latent, since the trace shows the strategy the user enacted rather than the unchosen alternatives or an explicit decision event. $P_t$ is therefore a tree over the constructs that remain observable, \emph{sequence}, \emph{for-each}, and \emph{while}, induced from the temporal order and the recurring patterns of $E_t$.

The three operators differ in the evidence that admits them. A \emph{sequence} node enumerates its child steps in temporal order. A \emph{for-each} node is admitted when the trace contains at least two aligned occurrences of the same activity pattern, differing mainly in the named artifact or entity being acted on (e.g., setting up and verifying sign-in for each account in \texttt{\{user\_1, user\_2\}}). A \emph{while} node is admitted when the repeated occurrences continue until a condition on the objective state is satisfied (e.g., implementing a change, rebuilding the website, and validating the fix until validation passes). Each repeated step must be grounded in the activities that realize it across the aligned occurrences, and a pattern that fails this test remains a sequence. Procedure-side validity constraints and repair steps are detailed in Appendix~\ref{app:structural-validity}.

\paragraph{Model reconciliation.}
The two models are internally consistent, yet they split the same activities differently. An objective decomposition can split one iterative unit across separate phases, and a procedure model can place a goal transition inside a flat \emph{sequence}. Reconciliation fuses $O_t$ and $P_t$ into one task model $M_t$ by fixing how each node's activities are divided among its children at each layer.

$M_t$ is expanded from the task objective downward. At each node, the control-flow operator comes from $P_t$ and the child objectives come from $O_t$. Where the two disagree, nodes are split, merged, or re-parented until objective scope and control flow agree, and Appendix~\ref{app:structural-validity} states the rule for each case. In \reffig{fig:method}(c), $P_t$ reads the repeated setup as one \emph{for-each} while $O_t$ splits it into a sub-objective per user. $M_t$ keeps the single loop and recovers the steps each model had lost on its own. Every node of $M_t$ then carries an objective and a control-flow operator, and every leaf stays grounded in $E_t$.

\subsection{Implementation}\label{sec:implementation}
All pipeline stages use gpt-5.4 at temperature 1.0. Direct generation also uses gpt-5.4; gpt-5.5 and claude-sonnet-5 serve as independent judges for intrinsic evaluation; and gpt-5-mini generates skills and executes held-out tasks in the extrinsic evaluation. Appendix~\ref{app:pipeline-prompts} lists the prompt template details. \reffig{fig:worked-example} shows the tasks and the task model induced from one recorded session.

%% file: sections/4_intrinsic_evaluation.tex
\begin{figure*}[!t]
    \centering
    \includegraphics[width=0.76\textwidth]{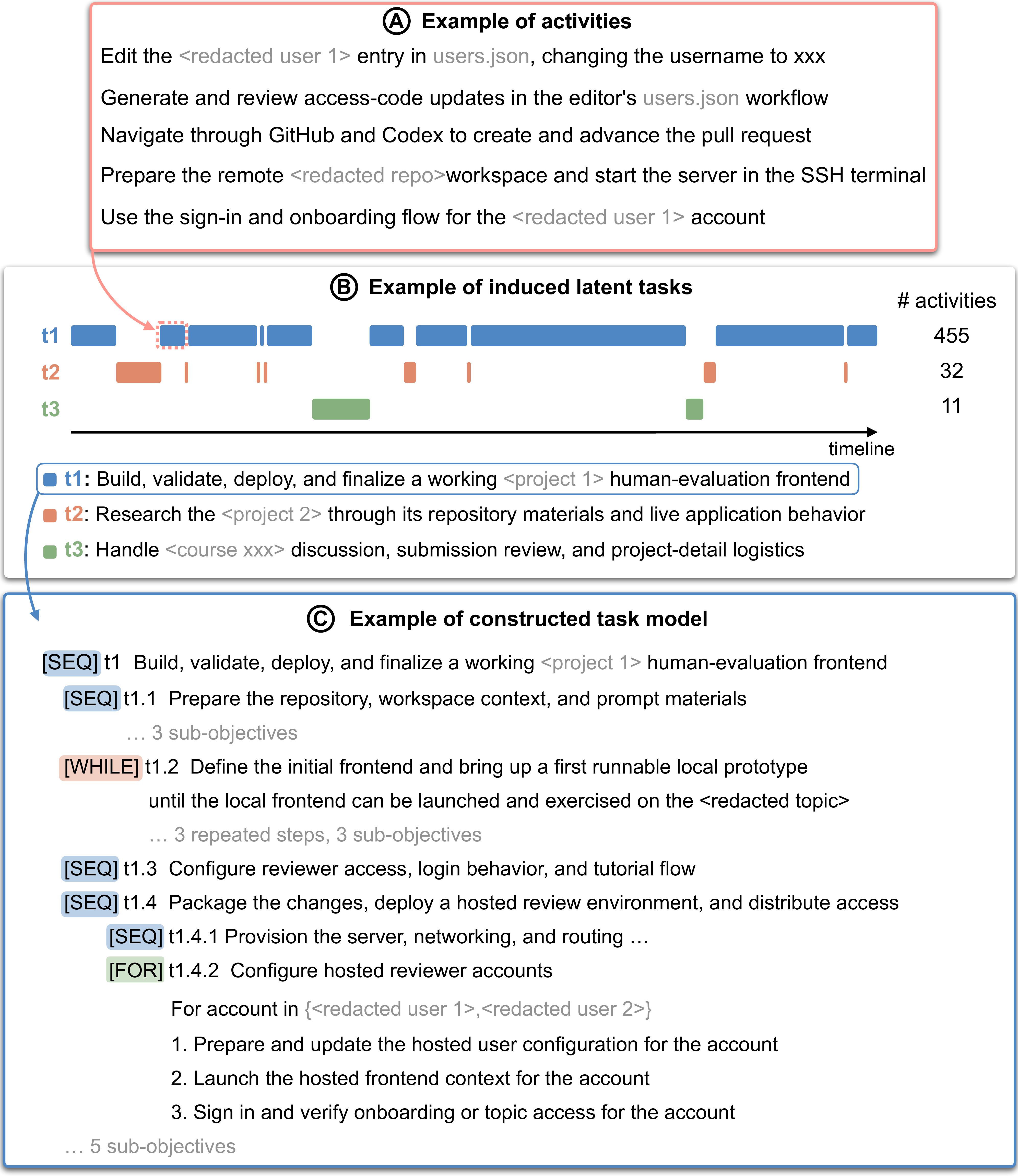}
    \caption{\textbf{Worked example of \textsc{TMI} applied to one recorded session}, which captures the construction of a web frontend for a human-evaluation study. \emph{(A)} Activities recovered by event grounding and segmentation (\S\ref{sec:activity-abstraction}). \emph{(B)} Latent tasks induced over the session (\S\ref{sec:latent-task-induction}). Three tasks are recovered with no task set given in advance, and none occupies a contiguous span. \emph{(C)} The task model constructed for $t_1$ (\S\ref{sec:task-model-construction}), with partial task model elided at $\cdots$. Every task model node carries an objective and a control-flow operator.
    }
    \label{fig:worked-example}
\end{figure*}

\paragraph{Datasets.}
We use two base datasets across the three intrinsic evaluations. The first, which we denote as HumanWork, is the human computer-use dataset from \citet{wang2025ai}. It comprises 38 recorded sessions spanning 15 tasks across five professional domains (data analysis, engineering, computation, writing, and design), with multiple human users per task and each session labeled with its ground-truth task identity. The sessions total 42.8 hours of recording and 48.7K raw events, averaging 68 minutes of recording and 1.3K events per session. The second is SkillsBench~\citep{li2026skillsbench}, comprising 86 software engineering tasks executed by coding agents across three harnesses (Claude Code, Codex, and Gemini CLI) and three skill conditions (Appendix~\ref{app:skillsbench_conversion}). We select the 15 tasks with the highest average action count, yielding 195 runs with 24.1K agent steps over 49.7 hours of execution. Each evaluation adapts these sources as described below.

\subsection{Latent Task Induction Robustness}
\label{sec:latent_task_robustness}
Real sessions interleave actions from many concurrent tasks, and a usable task model must recover each one without conflation. We measure how reliably the induction process separates interleaved actions into the correct latent tasks as task multiplicity and interleaving density grow.

\paragraph{Setup.}
To test robustness under varying degrees of task multiplicity and interleaving density, we construct synthetic multi-task trajectories from the HumanWork~\citep{wang2025ai} sessions, each of which records one task carried out end to end.
We sample $K$ sessions, cut each into $d$ contiguous segments, and shuffle the segments into one composite trajectory, so every task reappears at $d$ non-contiguous points. Larger $K$ means more concurrent tasks and larger $d$ means more frequent switching between them. We sweep $K$ from 2 to 15 and $d$ from 2 to 4, generating three independent trajectories per condition, and full construction details are in Appendix~\ref{app:synthetic}.

Each trajectory is processed through the full grounding and segmentation stage (\S\ref{sec:activity-abstraction}) followed by latent task induction (\S\ref{sec:latent-task-induction}). The system produces a predicted task count $\hat{K}$ and an assignment of activities to predicted tasks.
We assess task assignment with the Adjusted Rand Index~\citep{hubert1985comparing} ($\text{ARI}$), which measures pairwise agreement between predicted and ground-truth partitions of activities after correcting for chance (range $[-1, 1]$, higher is better) and  task count with mean absolute error $\text{MAE} = |\hat{K} - K|$, where lower values indicate more precise task enumeration. Both metrics are averaged across all trajectories per condition.

\paragraph{Induction stays robust under heavy interleaving.}
Table~\ref{tab:intrinsic_main} reports ARI and MAE aggregated over all task counts $K \in \{2,\ldots,15\}$ for each interleaving difficulty level. The system reaches 0.974 overall ARI, recovering non-contiguous task partitions even when activities are densely interwoven. MAE stays below 1 across all conditions, so the predicted task count tracks the true count as the number of concurrent tasks grows from 2 to 15. Appendix~\ref{app:error_analysis} presents the error analysis.

\begin{table}[t]
\centering
\small
\caption{Latent task induction performance by interleaving difficulty $d$ (number of segments per session), averaged over $K \in \{2,\ldots,15\}$ and three trajectories per condition. The bottom row reports the overall mean.}
\label{tab:intrinsic_main}
\begin{tabular}{c cc}
\toprule
$d$ & ARI $\uparrow$ & MAE $\downarrow$ \\
\midrule
2 & 0.980\,$\pm$\,0.024 & 0.29\,$\pm$\,0.34 \\
3 & 0.975\,$\pm$\,0.032 & 0.50\,$\pm$\,0.60 \\
4 & 0.968\,$\pm$\,0.029 & 0.64\,$\pm$\,0.61 \\
\midrule
Overall & 0.974\,$\pm$\,0.028 & 0.48\,$\pm$\,0.54 \\
\bottomrule
\end{tabular}
\end{table}

\subsection{Latent Task Induction Stability}
\label{sec:task_identity}
The induced latent task set should reflect the actual task being performed and remain stable across different execution strategies. Since induction never observes the task label, agreement with the assigned task indicates that induced tasks capture task identity rather than execution idiosyncrasies.

\paragraph{Metric.}
We evaluate on all 38 HumanWork~\citep{wang2025ai} sessions and on the SkillsBench runs whose task outcome is marked successful. For each run $i$, TMI observes only the trajectory and produces an induced latent task set $\mathcal{T}_i$. The dataset's ground-truth assigned task $y_i$ is withheld from induction and used only for evaluation. We give $\mathcal{T}_i$ and $y_i$ to the LLM judge, which returns a binary label indicating whether the induced task set correctly captures the assigned task. We report the average match rate across runs. To control for judge-family bias, we repeat the evaluation with judges from two model families, gpt-5.5 and claude-sonnet-5.

\paragraph{Induced tasks track task identity, not execution idiosyncrasies.}
Table~\ref{tab:task_identity} reports match rates for both datasets. On human sessions, the induced task set matches the assigned task in 94.74\% of runs, indicating that induction recovers the task identity without observing the dataset label. On successful SkillsBench runs, the overall match rate is 93.24\%, with similar rates across most agent harnesses and skill conditions. The induced latent task set changes with the underlying assigned task rather than merely reflecting idiosyncratic execution traces. The claude-sonnet-5 judge yields consistently high match rates, 89.47\% on human sessions and 98.65\% on SkillsBench, and does not favor task models induced from Claude Code trajectories, indicating that the result is not an artifact of a single judge family.

\begin{table}[t]
\centering
\footnotesize
\setlength{\tabcolsep}{3pt}
\caption{Latent task identity match rate (\%). Each judge compares the induced task set with the held-out assigned task. We report results from gpt-5.5 and claude-sonnet-5. SkillsBench includes only successful runs.}
\label{tab:task_identity}
\resizebox{\columnwidth}{!}{%
\begin{tabular}{@{}llrrr@{}}
\toprule
Dataset & Subset & $N$ & \multicolumn{2}{c}{Judge} \\
\cmidrule(lr){4-5}
 & & & gpt-5.5 & claude-sonnet-5 \\
\midrule
HumanWork & All & 38 & 94.74 & 89.47\phantom{0} \\
\midrule
SkillsBench & All & 74 & 93.24 & 98.65\phantom{0} \\
\addlinespace[2pt]
 & \textit{By agent} \\
 & \quad Claude Code & 23 & 95.65 & 100.00 \\
 & \quad Codex & 19 & 84.21 & 94.74\phantom{0} \\
 & \quad Gemini CLI & 32 & 96.88 & 100.00 \\
\addlinespace[2pt]
 & \textit{By skill} \\
 & \quad None & 21 & 95.24 & 95.24\phantom{0} \\
 & \quad Self-generated & 13 & 84.62 & 100.00 \\
 & \quad Curated & 40 & 95.00 & 100.00 \\
\bottomrule
\end{tabular}
}
\end{table}

\subsection{Task Model Fidelity}
\label{sec:structural_validity}
Downstream use relies on sub-goal decompositions that accurately capture task phases and procedure steps that faithfully describe the agent's execution. We assess the structural quality of the induced task model against human judgment and two baselines.

\paragraph{Baselines.}
We compare against two baselines. Workflow summary uses the workflow summarization toolkit from \citet{wang2025ai}, which produces a phase-based narrative of each observed session without a formal task model schema or control-flow operators. Direct generation prompts gpt-5.4 with the complete task model schema alongside the observed activity trace and asks the model to produce a task model in one pass. Both baselines use gpt-5.4 for generation. We further conduct ablation studies on the three components our method combines. \emph{w/o objective model} and \emph{w/o procedure model} score each model as it stands before reconciliation, on the dimensions that model defines. \emph{w/o reconciliation} keeps both models but induces them in a single joint pass and merges them, rather than resolving each under its own evidence and reconciling them afterwards.

\paragraph{Metrics.} We evaluate on all 38 HumanWork~\citep{wang2025ai} sessions. For human validation, we randomly sample 20 sessions and assign each to two independent annotators who score the same rubric dimensions (Appendix~\ref{app:human_agreement}).
We assess six rubric dimensions. Latent task recovery and objective coverage are rated on a 5-point Likert scale, and four per-node binary checks cover objective coherence, parent-child consistency, step description accuracy, and operator correctness. Full rubric definitions are in Appendix~\ref{app:fidelity_rubric}. The LLM judge receives the task instruction, activity trace, and induced task model. As in \S\ref{sec:task_identity}, we report scores produced by gpt-5.5 and claude-sonnet-5.

\paragraph{Task models stay faithful to procedures and observed objective decomposition.}
Table~\ref{tab:fidelity} reports results. Our method substantially outperforms both baselines on procedural fidelity, achieving 74.9\% step description accuracy versus 30.3\% baseline and 88.5\% operator correctness versus 52.7\% under the gpt-5.5 judge; the same ordering holds under claude-sonnet-5. On objective decomposition, our method performs comparably to direct generation, while yielding higher coverage under both judges and higher task recovery under gpt-5.5.

\paragraph{Neither model alone is sufficient, and reconciliation recovers what each one misses.}
The objective-only model achieves lower coherence and boundary-grounding scores because it overlooks procedural structure. When an action subsequence is repeated multiple times or until a condition is satisfied, the model often splits actions serving the same objective across multiple nodes. Conversely, the procedure-only model faithfully represents the observed actions but frequently misses their underlying purpose, causing transitions between objectives to be absorbed into a single flat sequence. Consequently, it achieves only 63.2\% description accuracy, compared with 74.9\% for the reconciled model. Although joint induction combines the two views in a single pass, it does so at the cost of granularity, producing only half as many nodes as our model. Its higher per-node scores reflect its coarser representation rather than more accurate structure, and its procedure steps are the least accurate among the three variants (Appendix~\ref{app:objective_rubric_full}). By contrast, reconciliation directly addresses the complementary failure modes of the two independently induced models. The procedure model corrects the boundaries of 64.5\% of objective nodes; conversely, evidence from the objective model corrects the boundaries of 21.9\% of procedure nodes and changes the control-flow operators of 0.6\%. \textbf{These results support our hypothesis that the objective and procedure models capture distinct yet complementary evidence about the same execution.}

\paragraph{The hierarchy holds over pervasive non-linear execution.}
Repair, exploration, and detours fill real sessions, with error correction present in 89\% of them and exploratory search in 87\% (Table~\ref{tab:nonlinear}). A flat step list cannot express these spans, and they are what the \emph{while} and \emph{for-each} operators exist to represent. The hierarchy induced over them stays intact, with no systematic drop in parent-child consistency. Step description accuracy on the hardest of them, 66.7\%, still exceeds the 30.3\% the strongest baseline reaches over all nodes. Boundary placement falls furthest, since repair and exploration supply no crisp deliverable to anchor it.

\begin{table}[t]
\centering
\footnotesize
\setlength{\tabcolsep}{4pt}
\caption{Non-linear execution in the human sessions and its effect on task-model fidelity. Sess.\ is the share of sessions in which the behavior appears and Ep.\ the number of episodes. P-Ch., Desc.\ and Bnd.\ are parent-child consistency, step description accuracy, and boundary grounding over the nodes each behavior dominates (\%, $\uparrow$) under the gpt-5.5 judge, against the clean-span rates in the first row. The labeling protocol and full results are in Appendix~\ref{app:nonlinear_fidelity}.}
\label{tab:nonlinear}
\begin{tabular}{@{}l rr rrr@{}}
\toprule
Behavior & Sess. & Ep. & P-Ch. & Desc. & Bnd. \\
\midrule
Clean spans & N/A & N/A & 92.2 & 79.9 & 79.0 \\
\midrule
Error correction & 89 & 143 & 94.5 & 66.7 & 59.6 \\
Exploratory search & 87 & \phantom{0}87 & 92.9 & 66.7 & 60.2 \\
Redundant repetition & 84 & 132 & 85.0 & 78.5 & 55.0 \\
Trial and error & 66 & \phantom{0}52 & 95.7 & 77.6 & 65.7 \\
Task switching & 61 & \phantom{0}61 & 93.0 & 71.1 & 67.4 \\
Backtracking or revision & 53 & \phantom{0}34 & 97.0 & 84.8 & 60.6 \\
\bottomrule
\end{tabular}
\end{table}

\begin{table}[t]
\centering
\footnotesize
\setlength{\tabcolsep}{2pt}
\caption{Task model fidelity. Task is latent task recovery; Cov. is objective coverage; Coh. is objective coherence; P-Ch. is parent-child consistency; Desc. is step description accuracy; Op. is operator correctness. Task and Cov. are 5-point Likert scores ($\uparrow$); remaining columns are binary pass rates (\%,$\uparrow$). Each ablation removes one component of our method, the objective model, the procedure model, or the reconciliation step where the two models are induced jointly in a single pass. Best per judge in bold.}
\label{tab:fidelity}
\begin{tabular}{@{}l c ccc cc@{}}
\toprule
 & & \multicolumn{3}{c}{Objective} & \multicolumn{2}{c}{Procedure} \\
\cmidrule(lr){3-5}\cmidrule(lr){6-7}
Model & Task & Cov. & Coh. & P-Ch. & Desc. & Op. \\
\midrule
\multicolumn{7}{@{}l}{\textit{gpt-5.5 judge}} \\
Workflow summary & 2.87 & 3.24 & 65.7 & 62.2 & 30.3 & N/A \\
Direct gen.        & 3.63 & 4.00 & 87.0 & 97.3 & 23.4 & 52.7 \\
Ours               & 3.71 & 4.34 & 85.7 & 92.6 & \textbf{74.9} & \textbf{88.5} \\
\quad w/o objective model & 3.37 & 3.76 & 77.9 & 85.1 & 63.2 & 84.1 \\
\quad w/o procedure model & \textbf{3.74} & 4.34 & 78.9 & 96.7 & N/A & N/A \\
\quad w/o reconciliation & 3.66 & \textbf{4.40} & \textbf{99.1} & \textbf{99.1} & 56.0 & 70.3 \\
\midrule
\multicolumn{7}{@{}l}{\textit{claude-sonnet-5 judge}} \\
Workflow summary & 2.46 & 2.71 & 93.2 & 92.8 & 68.6 & N/A \\
Direct gen.        & \textbf{3.53} & 3.18 & 85.6 & 75.3 & 47.8 & 78.3 \\
Ours               & 3.45 & 3.68 & 97.2 & \textbf{97.2} & \textbf{87.8} & \textbf{91.8} \\
\quad w/o objective model & 3.21 & 3.55 & 81.1 & 89.7 & 76.6 & 80.5 \\
\quad w/o procedure model & \textbf{3.76} & 3.74 & 90.4 & 95.4 & N/A & N/A \\
\quad w/o reconciliation & 3.53 & \textbf{3.97} & \textbf{100.0} & 96.9 & 87.3 & 87.0 \\
\bottomrule
\end{tabular}
\end{table}

%% file: sections/5_extrinsic_evaluation.tex
\label{sec:extrinsic}

A useful task model should transfer beyond the demonstration it was induced from. We test whether an induced task model is an effective source for generating reusable skills that improve a downstream agent's accuracy on held-out tasks.

\paragraph{Experiment Protocol.}
We test this by using the induced task model as the source of learning for agent skill generation on SkillLearnBench~\citep{zhong2026skilllearnbench}, a benchmark for continual learning methods for agent skill generation on real-world tasks. SkillLearnBench contains 20 task families, each grouping multiple tasks of a similar nature that can be addressed with similar strategies. For each task family we induce a task model from a single successful demonstration of one instance and pass it to the skill creator of Codex,\footnote{The skill creation component of the Codex CLI agent, \url{https://github.com/openai/codex}.} which synthesizes a reusable skill. The skill is scored against the SkillLearnBench rubric and then deployed on held-out instances of the same family.

We compare three sources of learning, each fed into the same skill creator. Raw demonstration passes the grounded trace directly, the workflow summary baseline~\citep{wang2025ai} passes a phase-based summary of the demonstration, and \emph{Ours} passes the task model. A \emph{No skill} condition bounds performance without any generated skill, and a \emph{Human curated} condition substitutes the expert-written skills released with the benchmark for a generated one. All skills are generated and all held-out instances executed with gpt-5-mini, so performance differences isolate the source of learning. We report the five SkillLearnBench metrics of skill coverage, executability, safety, agent trajectory alignment, and held-out task accuracy.

\paragraph{The task model is an effective representation for skill transfer.}
Skills generated from our task models are more executable and transfer better to held-out tasks than skills generated from raw demonstrations or workflow summaries. As shown in Table~\ref{tab:extrinsic}, our model improves executability from 59.35 to 67.65 and held-out accuracy from 14.29 to 18.57 over the strongest baseline, a 30\% relative accuracy gain. The expert-written skills score highest on skill coverage at 93.59 but reach 10.00 held-out accuracy, below every induced source, so skill coverage and held-out accuracy do not move together in this setting. We report the curated condition as a reference point for skill quality rather than as an upper bound on accuracy.

\begin{table}[t]
\centering
\footnotesize
\setlength{\tabcolsep}{3pt}
\caption{Extrinsic evaluation on SkillLearnBench. Each row is a source of learning fed into the same skill creator. Cov. is skill coverage; Exec. is executability; Safe. is safety; Align. is agent trajectory alignment; Acc. is held-out task accuracy. All values are percentages ($\uparrow$). No skill produces no generated artifact, so the skill-quality columns are not applicable. Human curated baseline judges the expert-written skills released with the original benchmark.}
\label{tab:extrinsic}
\begin{tabular}{@{}l ccc cc@{}}
\toprule
 & \multicolumn{3}{c}{Skill quality} & \multicolumn{2}{c}{Execution} \\
\cmidrule(lr){2-4}\cmidrule(lr){5-6}
Source of learning & Cov. & Exec. & Safe. & Align. & Acc. \\
\midrule
No skill                 & N/A & N/A & N/A & 59.20 & \phantom{0}8.57 \\
Human curated      & 93.59 & 63.80 & 90.50 & 56.59 & 10.00 \\
\midrule
Raw demonstration        & 52.46 & 53.49 & \textbf{92.52} & 63.15 & 11.43 \\
Workflow summary & 54.07 & 59.35 & 91.74 & 66.18 & 14.29 \\
Ours                     & \textbf{54.95} & \textbf{67.65} & 90.57 & \textbf{67.99} & \textbf{18.57} \\
\bottomrule
\end{tabular}
\end{table}

%% file: sections/6_related_works.tex
\paragraph{Inducing representation from behavior}

Plan recognition infers goals from observed actions but presupposes a plan library or domain theory~\citep{kautz1986generalized}, while process mining discovers procedural models from event logs that already contain typed activities and case identifiers~\citep{van2011process}. Recent computer-use trace work grounds pixels into semantic operations~\citep{shaw2023pixels,yang2023setofmark,you2024ferret,zheng2024gpt}, extracts action sequences or instructional steps~\citep{chen2024sharinganextractuseraction,zang2025ms4ui}, learns persistent user models~\citep{shaikh2025gum,shaikh2026learning}, pretrains agents on demonstrations~\citep{lu2025video,song2026watchlearnlearninguse}, or analyzes trajectories under a known root task~\citep{wang2024agent,wang2025ai}. We drop these assumptions, inducing semantic actions, latent tasks, objective hierarchies, and procedures jointly from raw screen-and-input traces.

\paragraph{Multitasking and interleaving}

Field and cognitive studies show that knowledge workers continuously interleave goals across working spheres and incur measurable costs when switching~\citep{czerwinski2004diary,gonzalez2004constant,mark2008cost,iqbal2007task,salvucci2009threaded}, and event segmentation theory frames activity as hierarchically organized boundaries inferred from goal and state changes~\citep{zacks2001event}. Work on grounding agent memory in inferred user intent further argues that latent intent must be modeled to make sense of interleaved activity~\citep{yang-etal-2026-grounding}. These findings ground our choice to treat task identity as latent and to allow non-contiguous task spans, so that interleaved activity is resolved during task induction rather than carried into the procedure model.

%% file: sections/7_conclusion.tex
We introduced \textsc{Task Model Induction}, which abstracts raw computer-use events into activities, discovers the latent tasks they realize, and reconciles separately induced objective and procedure models into one task model per task. It turns everyday computer-use activity traces into durable, auditable records of how work are carried out. Experiments show that this representation recovers interleaved execution faithfully and yields better downstream agent skills than raw traces or workflow summaries. TMI has the potential to facilitate the study of how work is carried out across domains and to make these records reusable as knowledge for both people and agents.

%% file: sections/9_limitation.tex
TMI operates on naturalistic computer-use traces that may contain personally identifiable information. Future deployments that apply this method to raw computer-use traces should consider privacy redaction of screenshots and keyboard events before induction, so that private and sensitive content does not propagate into induced artifacts distributed for downstream uses such as skill learning. Studying the effect of such redaction on the induction quality is left to future work.

%% file: sections/8_ethics.tex
This work uses three publicly available datasets, which are all released for research use. We use them as released and do not attempt to identify any individual user. The synthetic multi-task trajectories constructed for the robustness evaluation are assembled by merging segments of these existing public sessions; no new data collection involving human subjects was conducted. Skill generation and agent evaluation are performed on benchmark tasks with no access to private user data. We do not foresee direct harms from this work.

%% file: appendix/structural_validity.tex
\section{Structural Validity Constraints}
\label{app:structural-validity}

Both $O_t$ and $P_t$ are subject to formal validity constraints that induction must satisfy. For the objective tree, every activity in $E_t$ must appear under exactly one leaf node (completeness and non-overlap); a node spanning a single activity has reached the level of a local objective and must remain a leaf; and every internal node must state a desired outcome rather than an interface action. For the procedure tree, every activity in $E_t$ must appear in at least one procedure node's reference set, every operator must belong to the closed primitive set, every for-each node must bind its iteration variable to an explicitly enumerated collection, and every while node must state an objective-state exit condition. A loop body is an abstract template whose every step maps to the activity episodes it covers across all repetitions, so a loop is admitted only when grounded in recurring evidence rather than asserted. Constraint violations are identified by a deterministic validator and fed back as structured feedback, prompting targeted repair, after which a recovery pass re-examines flat sequences for repeated bodies that should have been folded into a for-each or while node.

\paragraph{Boundary placement in reconciliation.}
Reconciliation (\S\ref{sec:task-model-construction}) expands $M_t$ from the task objective, the root of $O_t$, which governs all of $E_t$. Expanding a node divides its governed subsequence among the node's children, and each division point is a \emph{boundary}, the position in $E_t$ where one child's subsequence ends and the next begins. A boundary that both models place is retained. Where $P_t$ reads a subsequence as a while or for-each operator but $O_t$ splits it into several sub-objectives, $M_t$ keeps the subsequence as one iterative phase and nests those sub-objectives as semantic refinements within it. Where $P_t$ reads it as a flat sequence, $M_t$ keeps the temporal order and adopts the child boundaries from $O_t$. Where neither model exposes internal structure yet the activities show aligned repetitions over named artifacts or repeated attempts under an unmet objective-state condition, the operator is inferred from the trace.

%% file: appendix/latent_task_induction_robustness.tex
\section{Latent Task Induction Robustness}
\label{app:synthetic}

\paragraph{Dataset statistics.}
The 38 recorded HumanWork sessions~\citep{wang2025ai} average 1{,}282 raw keyboard and mouse actions and 1.01 hours of active computer use per session. Active duration excludes idle intervals exceeding ten minutes between consecutive actions, treating such gaps as disengagement rather than active task work, and therefore falls below the 68 minutes of recording per session reported in Section~\ref{sec:experiment}.

\paragraph{Construction.}
Synthetic interleaving trajectories are constructed from these 38 sessions. For each task count $K \in \{2, 3, \ldots, 15\}$ and interleaving difficulty $d \in \{2, 3, 4\}$, we sample $K$ distinct tasks without replacement and assign each a single randomly selected session. The session for each task is then partitioned into $d$ contiguous segments by sampling $d - 1$ split points uniformly at random from positions satisfying the constraint that every resulting segment contains at least ten activities. This constraint prevents configurations in which a task occupies only a trivially short span, which would not reflect realistic interleaving patterns in naturalistic computer use. The $K \times d$ segments are then randomly permuted to form the composite trajectory; within each segment, the original event ordering is preserved. Three independent trajectories are generated per $(K, d)$ condition for a total of $14 \times 3 \times 3 = 126$ synthetic trajectories across all conditions. Table~\ref{tab:intrinsic_full} reports the full results.

The sampled tasks create substantial ambiguity at the application and domain levels. Across the 5{,}040 task pairs in the 126 trajectories, 96.7\% share an application and 88.9\% belong to the same domain. Even after excluding browsers and operating-system utilities, 70.4\% share an application. This overlap is particularly pronounced in the high-concurrency conditions, where up to 15 tasks must be separated across only five domains. Performance remains similar in trajectories containing same-domain task pairs (ARI 0.973; $n=112$) and those without them (ARI 0.984; $n=14$).

\begin{table*}[t]
\centering
\small
\caption{Full per-condition results for latent task induction robustness. We reports mean\,$\pm$\,std over three trajectories.}
\label{tab:intrinsic_full}
\begin{minipage}[t]{0.47\textwidth}
\centering
\begin{tabular}{cc cc}
\toprule
$K$ & $d$ & ARI $\uparrow$ & MAE $\downarrow$ \\
\midrule
  2 & 2 & 1.000$\pm$0.000 & 0.00$\pm$0.00 \\
   & 3 & 1.000$\pm$0.000 & 0.00$\pm$0.00 \\
   & 4 & 0.994$\pm$0.009 & 0.33$\pm$0.47 \\
  \addlinespace[2pt]
  3 & 2 & 1.000$\pm$0.000 & 0.00$\pm$0.00 \\
   & 3 & 1.000$\pm$0.000 & 0.00$\pm$0.00 \\
   & 4 & 0.902$\pm$0.046 & 0.67$\pm$0.47 \\
  \addlinespace[2pt]
  4 & 2 & 0.983$\pm$0.024 & 0.67$\pm$0.94 \\
   & 3 & 0.998$\pm$0.003 & 0.33$\pm$0.47 \\
   & 4 & 0.997$\pm$0.004 & 0.00$\pm$0.00 \\
  \addlinespace[2pt]
  5 & 2 & 1.000$\pm$0.000 & 0.00$\pm$0.00 \\
   & 3 & 1.000$\pm$0.000 & 0.00$\pm$0.00 \\
   & 4 & 0.972$\pm$0.040 & 0.33$\pm$0.47 \\
  \addlinespace[2pt]
  6 & 2 & 0.998$\pm$0.003 & 0.00$\pm$0.00 \\
   & 3 & 0.987$\pm$0.018 & 0.33$\pm$0.47 \\
   & 4 & 0.934$\pm$0.093 & 1.00$\pm$1.41 \\
  \addlinespace[2pt]
  7 & 2 & 1.000$\pm$0.000 & 0.00$\pm$0.00 \\
   & 3 & 1.000$\pm$0.000 & 0.67$\pm$0.47 \\
   & 4 & 1.000$\pm$0.000 & 0.00$\pm$0.00 \\
  \addlinespace[2pt]
  8 & 2 & 0.920$\pm$0.113 & 0.00$\pm$0.00 \\
   & 3 & 0.945$\pm$0.077 & 0.33$\pm$0.47 \\
   & 4 & 0.984$\pm$0.011 & 1.00$\pm$0.82 \\
\bottomrule
\end{tabular}
\end{minipage}\hfill
\begin{minipage}[t]{0.47\textwidth}
\centering
\begin{tabular}{cc cc}
\toprule
$K$ & $d$ & ARI $\uparrow$ & MAE $\downarrow$ \\
\midrule
  \addlinespace[2pt]
  9 & 2 & 0.972$\pm$0.040 & 0.00$\pm$0.00 \\
   & 3 & 0.997$\pm$0.005 & 0.33$\pm$0.47 \\
   & 4 & 0.948$\pm$0.059 & 1.00$\pm$0.00 \\
  \addlinespace[2pt]
  10 & 2 & 0.951$\pm$0.065 & 0.33$\pm$0.47 \\
   & 3 & 0.938$\pm$0.076 & 0.67$\pm$0.47 \\
   & 4 & 0.947$\pm$0.074 & 0.00$\pm$0.00 \\
  \addlinespace[2pt]
  11 & 2 & 0.960$\pm$0.056 & 0.33$\pm$0.47 \\
   & 3 & 0.997$\pm$0.004 & 0.33$\pm$0.47 \\
   & 4 & 0.966$\pm$0.013 & 2.00$\pm$1.41 \\
  \addlinespace[2pt]
  12 & 2 & 0.980$\pm$0.024 & 1.00$\pm$1.41 \\
   & 3 & 0.991$\pm$0.013 & 0.33$\pm$0.47 \\
   & 4 & 0.972$\pm$0.028 & 1.00$\pm$0.82 \\
  \addlinespace[2pt]
  13 & 2 & 0.990$\pm$0.010 & 0.67$\pm$0.47 \\
   & 3 & 0.923$\pm$0.044 & 2.33$\pm$1.25 \\
   & 4 & 0.999$\pm$0.002 & 0.00$\pm$0.00 \\
  \addlinespace[2pt]
  14 & 2 & 0.999$\pm$0.001 & 0.33$\pm$0.47 \\
   & 3 & 0.962$\pm$0.045 & 0.33$\pm$0.47 \\
   & 4 & 0.984$\pm$0.007 & 1.33$\pm$0.47 \\
  \addlinespace[2pt]
  15 & 2 & 0.962$\pm$0.033 & 0.67$\pm$0.47 \\
   & 3 & 0.917$\pm$0.081 & 1.00$\pm$0.82 \\
   & 4 & 0.952$\pm$0.035 & 0.33$\pm$0.47 \\
\bottomrule
\end{tabular}
\end{minipage}
\end{table*}

\subsection{Error Analysis}
\label{app:error_analysis}

We manually sample 1{,}000 predicted decision boundaries produced across all 126 synthetic trajectories and code their associated task labels. Since a boundary can have labels on both sides or arise from a multiway decision, this yields 1{,}107 labels. Table~\ref{tab:error_analysis} reports the distribution of label-level failure patterns; 87.1\% of labels are correctly assigned or outside the scope of any named failure type. The remaining 12.9\% split across four structural causes described below, along with potential mitigations.

\begin{table}[ht]
\centering
\small
\caption{Label-level failure distribution across 1{,}107 predicted task labels from 126 synthetic trajectories.}
\label{tab:error_analysis}
\begin{tabular}{lrr}
\toprule
Failure pattern & Labels & \% \\
\midrule
Correct / not failure-relevant & 871 & 87.1 \\
Subgoal promotion & 68 & 6.8 \\
Recorder-induced clustering & 25 & 2.5 \\
Cross-task workspace sharing & 24 & 2.4 \\
Others & 12 & 1.2 \\
\bottomrule
\end{tabular}
\end{table}

\paragraph{Subgoal promotion (6.8\%).}
A coherent phase within a benchmark task, such as data cleaning before analysis or asset export before presentation, is predicted as a separate root task, inflating $\hat{K}$. Among the 68 affected labels, 33.8\% involve real user navigation to personal activities that are unrelated to the assigned benchmark task; these cases are in principle unresolvable because the trajectory contains no task-specific context for the off-task behavior.

\paragraph{Recorder-induced clustering (2.5\%).}
The screen-recording tool used in the original data collection generates setup, screenshot, and teardown operations that are interleaved across multiple benchmark tasks and carry no task-specific context. These activities accumulate into spurious clusters that the system treats as independent tasks.

\paragraph{Cross-task workspace sharing (2.4\%).}
The dataset spans 15 tasks across five professional domains, so multiple tasks within the same domain share the same tools and output artifacts, for example two data analysis tasks both editing the same Jupyter notebook, or two design tasks both working in Figma. When such tasks are interleaved, their activity streams are very similar by tool or interface alone. The primary mitigation is to attend more to artifact-level unique identifiers and screen content, using both textual and visual signals to distinguish tasks that share an interaction surface.

%% file: appendix/skillsbench_trajectory_conversion.tex
\section{SkillsBench Trajectory Conversion}
\label{app:skillsbench_conversion}

\paragraph{Skill conditions.}
A \textit{skill} in SkillsBench is a reusable, human-authored procedural document that an agent may consult when approaching a task. Runs are executed with no supplementary skill (\textit{no-skill}), with a human-curated skill (\textit{skill}), or with a skill the agent generates for itself (\textit{self-generation}). 

\paragraph{Trajectory format.}
Coding agent trajectories from SkillsBench differ structurally from the low-level keyboard and mouse streams found in human computer-use recordings. Each step consists of an assistant message that issues one or more tool calls (e.g., \texttt{bash}, file reads, or web searches) followed by corresponding tool results. These operations are already semantic rather than raw input events, so they do not pass through the grounding and segmentation stage of our pipeline. Instead, we map each (tool call, tool result) pair directly to one activity in our schema, bypassing the low-level event grounding step.

\paragraph{Pairing strategy.}
Consecutive assistant turns can issue multiple tool calls in a single burst before any results arrive, a pattern common when Claude Code issues parallel independent commands. We apply a burst-aware last-$M$ assignment rule in which a burst of $N$ tool calls followed by only $M \leq N$ results maps the first $N - M$ calls to empty output (capturing write-style commands that produce no meaningful stdout) and the remaining $M$ calls are paired with results in FIFO order. When $M = N$ all calls are matched one-to-one. User-role messages and assistant text-only reasoning steps carry no tool calls and are dropped; orphaned tool-result steps with no preceding burst are also discarded.

\paragraph{Dataset statistics.}
Applying this conversion to the 195 selected SkillsBench runs (15 tasks, 13 runs per task) yields 9{,}459 activities in total, averaging 48.5 activities per run. The wide range reflects task heterogeneity across tasks spanning security patching, performance debugging, scientific computation, and multimedia processing.

%% file: appendix/task_model_fidelity_rubric.tex
\section{Task Model Fidelity Rubric}
\label{app:fidelity_rubric}

The LLM judge and human annotators use the same three rubrics below. Each evaluation call receives the task instruction, activity trace, and induced task model, then returns scores for one rubric group. The judge is instructed to use only evidence from the provided inputs and to return structured JSON matching the schema.

\nolinenumbers

\begin{tcolorbox}[
  enhanced, breakable,
  title={\textbf{Rubric 1 \;|\; Latent Task Recovery} (scored 1--5, $\uparrow$)},
  colback=blue!4, colframe=blue!45!black,
  fonttitle=\small\bfseries, fontupper=\small,
  top=4pt, bottom=4pt
]
Evaluates whether the root objective of the induced task model correctly identifies the latent task the agent was pursuing.

\medskip
\begin{tabular}{@{} r @{\hspace{6pt}} p{0.86\linewidth} @{}}
\textbf{5} & Correctly identifies the task intent, deliverable, and abstraction level. Specific enough to distinguish the task but not tied to incidental tools or local actions. \\[3pt]
\textbf{4} & Mostly correct, with a minor issue in specificity, abstraction, or a secondary deliverable. \\[3pt]
\textbf{3} & Partially correct. Captures the broad goal or domain but misses an important part of the task intent, deliverable, or scope. \\[3pt]
\textbf{2} & Weak match. Describes only a local phase, a tool action, or a vague activity rather than the full latent task. \\[3pt]
\textbf{1} & Incorrect or unsupported. Describes the wrong task, hallucinates intent, or is not grounded in the trace. \\
\end{tabular}
\end{tcolorbox}

\begin{tcolorbox}[
  enhanced, breakable,
  title={\textbf{Rubric 2 \;|\; Subgoal Decomposition Quality}},
  colback=green!4, colframe=green!45!black,
  fonttitle=\small\bfseries, fontupper=\small,
  top=4pt, bottom=4pt
]
\textbf{Coverage score} (1--5, $\uparrow$): evaluates whether the subgoals span all major phases of the observed task.

\medskip
\begin{tabular}{@{} r @{\hspace{6pt}} p{0.86\linewidth} @{}}
\textbf{5} & All major phases of the task are represented by subgoals. \\[3pt]
\textbf{4} & Most major phases are represented; only minor phases are missing or merged. \\[3pt]
\textbf{3} & Some important phases are represented, but at least one major phase is missing or poorly covered. \\[3pt]
\textbf{2} & Decomposition captures only a small part of the task or uses very coarse/noisy phases. \\[3pt]
\textbf{1} & Subgoals do not meaningfully cover the trace. \\
\end{tabular}

\medskip
\textbf{Binary checks} (per non-root objective node, $\uparrow$):

\smallskip
\begin{tabular}{@{} p{0.38\linewidth} @{\hspace{6pt}} p{0.47\linewidth} @{}}
Coherent subgoals & Each subgoal groups activities serving one recognizable intermediate objective. \\[3pt]
Parent-child consistent & Child subgoals collectively advance the parent objective. \\
\end{tabular}

\smallskip
The LLM judge additionally scores three diagnostic checks per node, sibling non-overlap, granularity appropriateness, and boundary grounding, defined and reported in Appendix~\ref{app:objective_rubric_full}.
\end{tcolorbox}

\begin{tcolorbox}[
  enhanced, breakable,
  title={\textbf{Rubric 3 \;|\; Procedure Fidelity}},
  colback=orange!5, colframe=orange!55!black,
  fonttitle=\small\bfseries, fontupper=\small,
  top=4pt, bottom=4pt
]
\textbf{Binary checks} (per procedure node):

\smallskip
\begin{tabular}{@{} p{0.42\linewidth} @{\hspace{6pt}} p{0.43\linewidth} @{}}
Operator correct ($\uparrow$) & The control-flow operator is supported by the trace. \\[3pt]
Step descriptions accurate ($\uparrow$) & Step descriptions faithfully summarize what occurred. \\[3pt]
No hallucination ($\uparrow$) & The node introduces no step or detail absent from the trace. \\
\end{tabular}

\medskip
\textbf{Operator decision rules}:

\smallskip
\begin{tabular}{@{} p{0.18\linewidth} @{\hspace{6pt}} p{0.67\linewidth} @{}}
\texttt{SEQ} & Steps occur in the stated temporal order. \\[2pt]
\texttt{FOR} & The trace shows the same operation applied to multiple items. \\[2pt]
\texttt{WHILE} & The trace shows repeated attempts or checks until a stopping condition. \\[2pt]
\multicolumn{2}{@{}p{\linewidth}@{}}{If an operator is plausible but not directly evidenced, mark \emph{operator correct} as false.} \\
\end{tabular}
\end{tcolorbox}

\ifdefined\linenumbers\linenumbers\fi

\subsection{Human Annotation Agreement}
\label{app:human_agreement}

Two annotators independently scored all 20 sampled sessions. Each scored the two session-level dimensions and a shared random sample of five objective nodes and five procedure nodes per session, yielding 995 paired binary judgments across the five checks per node. They agreed on 85\% of judgments (Cohen's $\kappa=0.48$; Gwet's $\mathrm{AC1}=0.79$). Agreement on procedure checks was $\kappa=0.54$. Objective checks are strongly skewed toward positive labels, for which Cohen's $\kappa$ is conservative; Gwet's $\mathrm{AC1}$ for these checks was 0.86. Table~\ref{tab:annotator_scores} reports the score underlying this agreement.

\begin{table}[t]
\centering
\footnotesize
\setlength{\tabcolsep}{3pt}
\caption{Rubric scores for our induced task models over the 20 annotated sessions. Task and Cov.\ are 5-point Likert scores ($\uparrow$); remaining columns are binary pass rates (\%, $\uparrow$) over the annotated nodes, with both LLM judges restricted to the same nodes. Both judges fall within the range spanned by the annotators on all four binary checks and score both Likert dimensions below the annotator mean, so the automatic evaluation does not inflate the reported levels.}
\label{tab:annotator_scores}
\setlength{\tabcolsep}{2pt}
\begin{tabular}{@{}l cc cccc@{}}
\toprule
Scorer & Task & Cov. & Coh. & P-Ch. & Desc. & Op. \\
\midrule
Annotator 1 & 4.20 & 4.85 & 98.0 & 100.0 & 93.0 & 93.0 \\
Annotator 2 & 3.85 & 4.10 & 87.9 & 93.9 & 70.0 & 87.0 \\
\midrule
gpt-5.5 judge & 3.75 & 4.35 & 88.9 & 94.9 & 70.0 & 91.0 \\
claude-sonnet-5 judge & 3.35 & 3.55 & 98.0 & 97.0 & 84.0 & 90.0 \\
\bottomrule
\end{tabular}
\end{table}

\subsection{Fidelity on Non-linear Execution}
\label{app:nonlinear_fidelity}

Naturalistic work contains repair, exploration, and detours that are not well described as a single forward procedure. We identify maximal spans of semantic actions exhibiting these behaviors, then associate a task-model node with a behavior when that behavior covers more than one third of its activity span. Automated episode discovery is verified by a human annotator, and unsupported episodes are discarded. Table~\ref{tab:nonlinear_prevalence} characterizes these episodes across the 38 sessions, and Table~\ref{tab:fidelity_diagnostics} reports their effect on task-model fidelity.

\begin{table}[t]
\centering
\small
\caption{Statistics on occurrence of non-linear execution behavior observed in the human sessions.}
\label{tab:nonlinear_prevalence}
\resizebox{\columnwidth}{!}{%
\begin{tabular}{lrrr}
\toprule
Behavior & Episodes & Sessions & Median length \\
\midrule
Error correction & 143 & 89\% & 9 \\
Redundant repetition & 132 & 84\% & 13 \\
Exploratory search & 87 & 87\% & 12 \\
Task switching & 61 & 61\% & 5 \\
Trial and error & 52 & 66\% & 15 \\
Backtracking or revision & 34 & 53\% & 14.5 \\
\bottomrule
\end{tabular}
}
\end{table}

\begin{table}[t]
\centering
\scriptsize
\caption{Task-model fidelity on non-linear execution. A node is associated with a behavior when it occupies more than one third of the node's span. Values are binary pass rates (\%) pooled over nodes, so the All nodes row differs marginally from the per-session means in Table~\ref{tab:fidelity}.}
\label{tab:fidelity_diagnostics}
\begin{tabular}{lrrr}
\toprule
Behavior & Coherent & P-Ch. & Boundary \\
\midrule
All nodes & 85.7 & 92.6 & 66.7 \\
Clean spans & 87.4 & 92.2 & 79.0 \\
Error correction & 82.6 & 94.5 & 59.6 \\
Exploratory search & 78.6 & 92.9 & 60.2 \\
Redundant repetition & 86.0 & 85.0 & 55.0 \\
Trial and error & 88.6 & 95.7 & 65.7 \\
Task switching & 79.1 & 93.0 & 67.4 \\
Backtracking or revision & 97.0 & 97.0 & 60.6 \\
\bottomrule
\end{tabular}

\medskip
\begin{tabular}{lrr}
\toprule
Behavior & Step desc. & No halluc. \\
\midrule
All nodes & 75.0 & 76.3 \\
Clean spans & 79.9 & 81.7 \\
Error correction & 66.7 & 64.3 \\
Exploratory search & 66.7 & 68.5 \\
Redundant repetition & 78.5 & 79.4 \\
Trial and error & 77.6 & 80.3 \\
Task switching & 71.1 & 73.3 \\
Backtracking or revision & 84.8 & 87.9 \\
\bottomrule
\end{tabular}
\end{table}

\subsection{Error Propagation across the Pipeline}
\label{app:error_propagation}

The pipeline progressively converts low-level observations into a unified task model, so an early error can affect later structural inferences. We trace failed nodes to their earliest originating stage using conditional failure statistics from 1{,}535 per-node fidelity judgments and a manual coding of 120 sampled failures. Grounding errors, primarily retained recorder operations and off-task browsing, affect approximately 9\% of objective nodes and 5\% of procedure nodes. Segmentation errors, chiefly activities that merge two intents, affect approximately 6\% of objective nodes and persist because later stages do not re-segment the trace.

Latent-task induction is stable, with 37 of 38 sessions inheriting no error from this stage, while the single failed induction loses most nodes in that session. Objective-model errors affect approximately 4\% of nodes but can propagate vertically, with 83\% of children failing under a mis-scoped parent compared with 24\% counterpart. Procedure-model errors are more localized, principally inaccurate leaf descriptions. Reconciliation has the largest effect on end-to-end fidelity, since removing it doubles procedure hallucination, while misplaced boundaries remain its main residual error, affecting approximately 12\% of nodes. These patterns distinguish errors that change the recovered task structure from those confined to local descriptions.

\subsection{Stability across Induction Runs}
\label{app:induction_stability}

We run the complete pipeline three times, resampling every LLM stage, in addition to the primary induction over the 38 human sessions. Latent task identity match is $91.2 \pm 4.0$ under gpt-5.5 and $90.4 \pm 4.0$ under claude-sonnet-5. Table~\ref{tab:induction_stability} shows corresponding variation in task-model fidelity.

\begin{table}[t]
\centering
\footnotesize
\setlength{\tabcolsep}{3pt}
\caption{Task-model fidelity across three independent induction runs over the 38 human sessions. Rows follow the columns of Table~\ref{tab:fidelity}; repeated-induction results are mean $\pm$ standard deviation.}
\label{tab:induction_stability}
\begin{tabular}{@{}l cc@{}}
\toprule
Dimension & gpt-5.5 & claude-sonnet-5 \\
\midrule
Task & 3.70$\pm$0.05 & 3.35$\pm$0.13 \\
Cov. & 4.41$\pm$0.06 & 3.65$\pm$0.10 \\
\addlinespace[2pt]
Coh. & 92.27$\pm$5.69 & 97.17$\pm$0.06 \\
P-Ch. & 95.83$\pm$2.90 & 98.00$\pm$1.47 \\
Desc. & 69.07$\pm$5.49 & 90.10$\pm$2.04 \\
Op. & 92.27$\pm$3.28 & 94.77$\pm$2.71 \\
\bottomrule
\end{tabular}
\end{table}

%% file: appendix/objective_rubric_full_results.tex
\section{Full Objective Rubric Results}
\label{app:objective_rubric_full}

Beyond the coherence and parent-child consistency checks reported in Table~\ref{tab:fidelity}, the LLM judge scores three further binary checks per objective node. Sibling non-overlap requires that sibling subgoals do not claim the same activities or restate one another; granularity appropriateness requires that a subgoal sit at a meaningful intermediate level, neither restating the parent nor describing a single interface action; boundary grounding requires that the activity span of a subgoal start and end at boundaries supported by the trace. Table~\ref{tab:objective_full} reports all five checks, together with the session-level objective coverage score from Table~\ref{tab:fidelity}, the mean number of judged objective nodes per session, the rate of nodes passing all five checks jointly, and the mean number of such fully valid nodes per session.

\begin{table*}[ht]
\centering
\footnotesize
\caption{Full objective rubric results. Cov.\ is the objective coverage score from Table~\ref{tab:fidelity} (5-point Likert, $\uparrow$); Nodes is the mean number of judged objective nodes per session. Coh.\ is subgoal coherence; NOv.\ is sibling non-overlap; Gran.\ is granularity appropriateness; P-Ch.\ is parent-child consistency; Bnd.\ is boundary grounding; All 5 is the rate of nodes passing all five checks. All check columns are per-node pass rates (\%, $\uparrow$). /Sess.\ is the mean number of fully valid nodes per session ($\uparrow$). The claude-sonnet-5 judge covers 35 of 38 sessions for the workflow baseline. Best per judge in bold.}
\label{tab:objective_full}
\begin{tabular}{@{}l rr rrrrr r r@{}}
\toprule
Model & Cov. & Nodes & Coh. & NOv. & Gran. & P-Ch. & Bnd. & All 5 & /Sess. \\
\midrule
\multicolumn{10}{@{}l}{\textit{gpt-5.5 judge}} \\
Workflow summary & 3.24 & 106.7 & 65.7 & 4.9 & 18.4 & 62.2 & 64.3 & 0.9 & 0.9 \\
Direct gen.        & 4.00 & 3.8 & 87.0 & 58.2 & \textbf{96.6} & 97.3 & 37.0 & 27.4 & 1.1 \\
Ours               & 4.34 & 19.5 & 85.7 & \textbf{94.9} & 86.5 & 92.6 & 66.7 & 59.0 & \textbf{11.5} \\
\quad w/o reconciliation & \textbf{4.40} & 8.6 & \textbf{99.1} & 81.2 & 95.4 & \textbf{99.1} & \textbf{80.0} & \textbf{70.2} & 6.0 \\
\midrule
\multicolumn{10}{@{}l}{\textit{claude-sonnet-5 judge}} \\
Workflow summary & 2.71 & 105.2 & 93.2 & 20.1 & 29.5 & 92.8 & 60.7 & 8.4 & 8.9 \\
Direct gen.        & 3.18 & 3.8 & 85.6 & 21.9 & 76.0 & 75.3 & 31.5 & 10.3 & 0.4 \\
Ours               & 3.68 & 19.5 & 97.2 & \textbf{94.3} & \textbf{92.6} & \textbf{97.2} & \textbf{82.7} & \textbf{78.1} & \textbf{15.2} \\
\quad w/o reconciliation & \textbf{3.97} & 8.6 & \textbf{100.0} & 73.8 & 89.8 & 96.9 & 78.5 & 59.7 & 5.1 \\
\bottomrule
\end{tabular}
\end{table*}

\paragraph{Per-node rates and decomposition granularity.}
The rates in Table~\ref{tab:fidelity} are averaged over the nodes each method produces, so they do not reflect how much structure a model recovers. Direct generation induces 3.8 objective nodes per session and joint induction 8.6, against 19.5 for our method, and coarser nodes face easier consistency checks. The model without reconciliation leads only on the two checks least sensitive to granularity and trails our method on sibling non-overlap under both judges, while the workflow baseline shows that volume alone does not help, as its flat lists of over 100 steps score lowest on granularity and overlap. Weighing validity and richness together, our method yields about twice as many fully valid subgoal nodes per session as joint induction, an order of magnitude more than direct generation, and the highest joint pass rate under the claude-sonnet-5 judge.

\paragraph{Where consistency failures occur.}
Objective and procedure nodes share identifiers in our unified model, which allows a node-level cross-tabulation of the two rubric groups. Objective nodes of our method that fail the coherence check carry an inaccurate step description at a rate of 0.50 against 0.21 for passing nodes under the gpt-5.5 judge, and at 0.43 against 0.12 under claude-sonnet-5. Inspecting the failing nodes shows that they concentrate on execution-shaped stretches of work, chiefly iterative repair and verification loops, re-establishment of working context after switches, and auxiliary setup such as authentication. Because reconciliation requires every observed activity to be covered by an objective node, these stretches surface as subgoals whose intent is defined by the course of execution rather than by a crisp deliverable, which the coherence check penalizes. One-pass induction absorbs the same stretches into broader nodes rather than surfacing them, which spares its consistency rates, yet under both judges it produces half or fewer fully valid objective nodes per session and recovers procedures less faithfully (Table~\ref{tab:fidelity}). Its higher consistency rates reflect what its decompositions absorb, not better objective modeling.

%% file: appendix/pipeline_prompts.tex
\section{Implementation Details and Pipeline Prompt Templates}
\label{app:pipeline-prompts}

This appendix lists the fixed instruction templates used by each LLM call in the pipeline. The two action-grounding prompts (Figure~\ref{lst:prompt-action-grounding}) are used by the vision-language grounding step described in Section~\ref{sec:activity-abstraction}. The backward semantic-action segmentation prompt (Figure~\ref{lst:prompt-semantic-actions}) and the activity segmentation prompt (Figure~\ref{lst:prompt-activity-segmentation}) are both used by the two segmentation passes described in the same section. The latent task discovery and consolidation prompts (Figures~\ref{lst:prompt-task-thread-discovery} and~\ref{lst:prompt-task-thread-consolidation}) are used by the two-phase latent task induction in Section~\ref{sec:latent-task-induction}. The objective-model, procedure-model, and reconciliation prompts (Figures~\ref{lst:prompt-objective-model}, \ref{lst:prompt-procedure-model}, and~\ref{lst:prompt-reconciliation}) are used by the objective, procedure, and reconciliation steps in Section~\ref{sec:task-model-construction}.

\begin{figure*}[!ht]
\begin{prompt}
GOAL_SYSTEM_PROMPT = """
Infer the immediate intent of one computer action.

Use the action string, the screenshot captured at the action moment, and the optional after screenshot. The before/action screenshot is primary evidence; use the after screenshot only to disambiguate what changed.

Return one concise sentence for the `goal` field.

Rules:
- Describe the local UI operation, not the user's broader task.
- Prefer concrete visible targets: button names, menu items, fields, files, tabs, cells, links, commands, or text snippets.
- Include the action verb when it matters, such as click, drag, type, select, open, close, scroll, or submit.
- Do not invent hidden motivations or off-screen content.
- If the visible evidence is insufficient, return the best grounded statement and mark uncertain details as "not sure".
"""

CONTEXT_SYSTEM_PROMPT = """
Ground one computer action in visible UI context.

Use the action string, the screenshot captured at the action moment, any zoomed-in crops, and the optional after screenshot. Zoom crops are centered on the action coordinates; a red outline or marker indicates the likely target region.

Return:
- `active_application`: application name plus visible window, page, document, file, or tab title when readable.
- `visual_content`: the specific visible artifact the action is aimed at or the user's eyes are likely focused on.

Rules:
- Do not output the goal; only output application/context fields.
- For `active_application`, prefer formats like "Google Chrome - Page title", "VS Code - filename.py", "Terminal - shell session", or "not sure".
- For `visual_content`, name the exact visible control/content region when possible: button, menu item, field, selected text, file row, cell, chart, code line, terminal command, tab, or document section.
- Ground every detail in visible text, recognizable UI, the action coordinate, or the before/after change.
- If a field is not clearly visible, return "not sure" for that field.
"""
\end{prompt}
\caption{Action grounding prompts for immediate intent and visible UI context.}
\label{lst:prompt-action-grounding}
\end{figure*}

\begin{figure*}[!ht]
\begin{prompt}
BACKWARD_SEMANTIC_ACTION_SEGMENT_PROMPT = """
You are analyzing a user's computer workflow by looking at actions in REVERSE order from the end of the session backward.

KEY INSIGHT: later outcomes help explain earlier low-level actions.

Each segment should be one candidate atom semantic action.

{semantic_action_definition}

=== WHAT HAPPENS AFTER THESE ACTIONS ===
{future_context}

=== ACTIONS TO ANALYZE (chronological order, index 0 = earliest) ===
{actions_list}

=== TASK ===
Segment these low-level actions into atom semantic actions.

For each group output:
1. semantic_action: one concise sentence describing the intentional operation

Semantic-action rules:
- Use semantic, operation-level language.
- Prefer the immediate operation over the broader task objective.
- Keep concrete apps, clicks, typing, commands, URLs, files, and navigation out of semantic_action unless essential.
- Avoid semantic actions that start with purely mechanical verbs like click, scroll, focus, move, hover, drag, or wait unless that operation is itself the meaningful user action.
- Do not skip failed attempts or corrections; include them with the operation they are trying to complete when intent is unchanged.

Coverage rules:
- Groups must be consecutive indices with no gaps, overlaps, or reorder.
- Every index from 0 to {max_idx} must appear exactly once.
- Split when the range contains multiple intentional operations.
- Keep together repeated low-level interaction needed to finish the same operation.

Output ONLY valid JSON:
{
  "groups": [
    {"start_idx": <int>, "end_idx": <int>, "semantic_action": "<atom semantic action>"}
  ]
}

List groups in REVERSE chronological order, latest group first. start_idx and end_idx are inclusive.
"""
\end{prompt}
\caption{Backward semantic-action segmentation prompt.}
\label{lst:prompt-semantic-actions}
\end{figure*}

\begin{figure*}[!ht]
\begin{prompt}
SEGMENTATION_PROMPT = """
You segment chronological atom semantic actions into activities.

{definition}

=== WHAT HAPPENED BEFORE THIS BATCH ===
{prior_context}

=== SEMANTIC ACTIONS TO SEGMENT (chronological order, index 0 = earliest in this batch) ===
{actions_list}

=== TASK ===
Partition the current batch into contiguous activities.

For each segment output:
1. start_idx and end_idx, inclusive, using the batch-local indices.
2. objective: one concise, self-contained phrase/sentence naming the intended local outcome or intermediate state. Include the concrete target, artifact, person, project, channel, file, URL, or app needed to interpret the activity. Do not list the procedure.
3. additional_context: one to three concise sentences with the observed procedure and concrete evidence needed to understand that objective.

Coverage rules:
- Every index from 0 to {max_idx} must appear exactly once.
- Segments must be consecutive with no gaps, overlaps, or reordering.
- Keep scaffolding actions with the objective they enable when evidence supports it.
- If an activity appears to continue across a batch boundary, produce the best segment inside this batch; a later merge pass will join adjacent segments.

Output ONLY valid JSON:
{
  "segments": [
    {"start_idx": <int>, "end_idx": <int>, "objective": "<activity>", "additional_context": "<concise evidence-grounded context>"}
  ]
}
"""
\end{prompt}
\caption{Activity segmentation prompt over semantic actions.}
\label{lst:prompt-activity-segmentation}
\end{figure*}

\begin{figure*}[!ht]
\begin{prompt}
ROOT_THREAD_DISCOVERY_PROMPT = """
You are building a task thread forest from chronologically ordered LEAF latent tasks.

Each leaf is already a local task. Your job is to attach each leaf to a durable ROOT thread.

Core mental model:
- Ask: "Which long-running top-level objective is this leaf advancing right now?"
- Roots represent durable objectives / deliverables, not contiguous time blocks.
- Leaves may interleave across roots.
- A root can pause and later resume.

Critical rules:
- Objective continuity beats adjacency.
- Two adjacent leaves can belong to different roots.
- Two distant leaves can belong to the same root.
- Different apps do NOT imply different roots.
- Interruptions do NOT imply different roots.
- Create a NEW root only when a genuinely new durable objective appears.
- Prefer a small number of strong roots over many near-duplicates.
- Communication leaves belong to the root defined by the SUBJECT of the message, not to a generic communication root.
- Setup, debugging, repo inspection, and environment preparation should stay under the same root as the later deliverable if they are clearly in service of that deliverable.

Available existing roots:
{existing_roots}

Most recent assigned leaves before this batch:
{recent_context}

Current leaves to assign:
{leaf_batch}

Task:
1. Reuse an existing root whenever the leaf advances the same durable objective / deliverable.
2. Create a new root only when needed.
3. New roots created inside this batch can be referenced by later leaves in the same batch.
4. After assigning leaves, update each touched root's label/objective/summary/last_update/anchor so future batches can judge fit.

Root update rules:
- summary is at most two sentences describing the durable thread so far.
- last_update is exactly one concise sentence describing the latest assigned leaf or leaves.
- anchor is a minimal concise list of stable identifiers for matching future work: project names, repos, datasets, products, people, files, or systems. Normalize aliases when they clearly refer to the same project; for example, two different names for the same codebase or product should share one anchor entry.
- Do not let anchor grow into a keyword dump. Prefer 1-5 meaningful identifiers.

Output ONLY valid JSON with `new_roots`, `assignments`, and `root_updates`.
"""
\end{prompt}
\caption{Latent task discovery prompt.}
\label{lst:prompt-task-thread-discovery}
\end{figure*}

\begin{figure*}[!ht]
\begin{prompt}
ROOT_THREAD_CONSOLIDATION_PROMPT = """
You are consolidating provisional durable root threads into the final task thread forest.

Each provisional root was discovered from chronological leaves. Some provisional roots may actually belong to the SAME durable objective and should be merged.

Core rules:
- Merge provisional roots if they advance the same long-running deliverable / objective, even if they are far apart in time, use different apps, or are interrupted.
- Keep roots separate if they represent genuinely different durable objectives.
- Objective continuity beats adjacency.
- Tiny opportunistic one-off roots may be absorbed into a nearby substantive root if they do not establish an independent durable objective.
- Prefer a compact set of strong canonical roots.
- Early setup/debugging/investigation roots should be merged into the later product root when they clearly enable that same deliverable.
- Communication-heavy provisional roots should be merged based on what the messages are ABOUT, not merely because they happen in the same messaging tool.

Provisional roots:
{provisional_roots}

Output ONLY valid JSON:
{
  "canonical_roots": [
    {
      "canonical_root_id": "C1",
      "label": "<short human-readable root label>",
      "objective": "<durable top-level objective>",
      "deliverable": "<artifact/state this root advances>",
      "success_criteria": "<observable completion criteria>",
      "member_root_ids": ["R001", "R004"]
    }
  ]
}
"""
\end{prompt}
\caption{Latent task consolidation prompt.}
\label{lst:prompt-task-thread-consolidation}
\end{figure*}

\begin{figure*}[!ht]
\begin{prompt}
GENERATION_PROMPT_TEXT = """
You induce a hierarchical objective model from task activity observations using computational thinking and recursive decomposition.

The input contains activity segments describing WHAT A USER DID. Your job is to abstract over those actions and recover the hierarchy of SUB-GOALS they were pursuing -- the recursive decomposition of the task into what needs to be accomplished at each level.
Return only a valid JSON object.

Objectives follow the computational thinking paradigm of recursive decomposition:
- Each node states a SUB-GOAL: what needs to be accomplished at this level to advance the parent goal.
- Write as a goal to be achieved -- not as a low-level procedure and not as a passive state predicate.
- Objectives must be TOOL-AGNOSTIC and USER-AGNOSTIC.
- Save procedural and evidential details for the `summary` field, NOT the `objective` field.

Granularity rules:
- A child node must represent a sub-outcome that is a necessary precondition or component of the parent outcome.
- If the input already represents one coherent atomic success state, use "decomposition": [] instead of inventing procedural children.
- A node covering exactly one activity/subgoal segment must not have decomposition.

Required output schema, recursively:
{
  "id": "<stable hierarchical id, e.g. C1 or C1.1>",
  "objective": "<sub-goal: what needs to be accomplished at this level, tool-agnostic and user-agnostic>",
  "summary": "<brief evidence-grounded summary; may reference specific tools, files, or actions observed>",
  "subgoal_segments": ["<single integer id such as 16 or closed integer range string such as 16-23>"],
  "decomposition": [<child nodes with the same schema> or <empty list if no further decomposition is needed>]
}
"""
\end{prompt}
\caption{Objective-model induction prompt.}
\label{lst:prompt-objective-model}
\end{figure*}

\begin{figure*}[!ht]
\begin{prompt}
PROCEDURE_GENERATION_PROMPT_TEXT = """
You induce a procedure model from a task-thread objective JSON by applying the Structured Programming Theorem (Bohm-Jacopini, 1966).
Return only a valid JSON object.

The Structured Programming Theorem motivates three control constructs:
1. Sequence -- steps executed one after another in order.
2. Selection -- a choice between mutually exclusive alternative paths.
3. Iteration -- a body repeated either over a named collection (for-each) or until a condition holds (while).

The trace records only the path enacted, not unchosen alternatives, so selection is not represented in the output. Decompose the observed activity trace into a tree using the three observable operators below.

The three operators are the complete and closed vocabulary:
- `SEQ`: steps in fixed order with no repetition and no branching.
- `FOR`: the same procedure body is applied to each member of a named, enumerable collection.
- `WHILE`: a body is repeated until an observable objective-state condition is satisfied.

FOR and WHILE bodies are ABSTRACT TEMPLATES -- they describe what happens per item / per pass using named steps with `name`, `description`, and `activity_refs`. Do NOT place activity_id leaves inside a FOR or WHILE body.

Coverage rules:
- Every activity episode in the input must appear in at least one node's `activity_refs` or as an inline `activity_id` leaf.
- Composite nodes cover the union of their children's episodes.
- Prefer one primary owning node per episode.

Required output schema:
{procedure_output_schema_text()}
"""
\end{prompt}
\caption{Procedure-model induction prompt.}
\label{lst:prompt-procedure-model}
\end{figure*}

\begin{figure*}[!ht]
\begin{prompt}
RECONCILIATION_GENERATION_PROMPT = """
You produce a unified task model by reconciling independently-induced objective and procedure models for the same activity trace. Return only a valid JSON object.

What you produce:
A unified tree where each node has two layers:
- Objective layer: a domain-specific program in natural language. Captures domain invariants: required outcomes, correctness constraints, and orderings that hold regardless of who executes the task or what specific inputs are used.
- Procedure layer: the faithful record of how the work was actually carried out in the observed trace, including failures and corrections.

Both layers are determined jointly from both input models. Neither is authoritative alone.

Inputs:
- `source`: task-thread JSON with an `activities` list.
- `objective_model`: hierarchical objective model.
- `procedure_model`: control-flow procedure model.

Reconciliation:
- Both models draw the same boundary: honor it.
- Procedure shows WHILE or FOR across a range: strong structural signal; that range is one iterative phase.
- Procedure shows only a flat SEQ: weak structural signal; defer to the objective model's semantic clusters.
- Objective shows a clear semantic phase shift: supports a new boundary even when the procedure model draws a continuous SEQ.
- Neither model shows structure for a range: scan the source `activities` for FOR or WHILE patterns.

Coverage and ID rules:
- Every source activity must appear in at least one node's `activity_refs`.
- A node's `activity_refs` is the union of its children's `activity_refs` or its body-step `activity_refs`.
- Use compact ranges: `activity_NNNN` or `activity_NNNN-activity_MMMM`.
- Root id is the task-thread id; children are numbered sequentially.

Required output schema:
{unified_schema_text()}
"""
\end{prompt}
\caption{Bidirectional objective/procedure reconciliation prompt.}
\label{lst:prompt-reconciliation}
\end{figure*}